\documentclass[letterpaper, 10 pt, conference, twocolumn]{ieeeconf}  
\usepackage{amsmath,amsfonts,amssymb}
\usepackage{algorithmic}
\usepackage{algorithm}
\usepackage[caption=false,font=normalsize,labelfont=sf,textfont=sf]{subfig}
\usepackage{textcomp}
\usepackage{stfloats}
\usepackage{url}
\usepackage{verbatim}
\usepackage{graphicx}
\usepackage{cite}
\usepackage{tabularx}
\usepackage{textcomp}
\usepackage{xcolor}
\usepackage{siunitx}
\usepackage{makecell}
\usepackage{comment}
\usepackage{array,multirow}
 \usepackage{float}
\usepackage{booktabs}
\usepackage{myMacros}
\usepackage{caption}
\usepackage{xcolor}
\usepackage{siunitx}  
\usepackage{hyperref}
\usepackage{nicefrac}
\usepackage{verbatim}
\IEEEoverridecommandlockouts                              

\title{\LARGE \bf
Learning-based NLOS Detection and Uncertainty Prediction of \\ GNSS Observations with Transformer-Enhanced LSTM Network
}

\author{Haoming Zhang, Zhanxin Wang and Heike Vallery
\thanks{This work was supported by the German Federal Ministry of Economic Affairs and Climate Action (BMWK) under Project 50NA2103 (Firefly).}
\thanks{All authors are with the Institute of Automatic Control, Faculty of Mechanical Engineering, RWTH Aachen University, 52074 Aachen, Germany. Heike Vallery is also with the Department of BioMechanical Engineering, Delft University of Technology, and with the Department for Rehabilitation Medicine, Erasmus MC, Rotterdam, The Netherlands.}
\thanks{Corresponding author: h.zhang@irt.rwth-aachen.de}
}

\begin{document}
\maketitle
\thispagestyle{empty}
\pagestyle{empty}

\begin{abstract}
The global navigation satellite systems (GNSS) play a vital role in transport systems for accurate and consistent vehicle localization. However, GNSS observations can be distorted due to multipath effects and non-line-of-sight (NLOS) receptions in challenging environments such as urban canyons. In such cases, traditional methods to classify and exclude faulty GNSS observations may fail, leading to unreliable state estimation and unsafe system operations. This work proposes a deep-learning-based method to detect NLOS receptions and predict GNSS pseudorange errors by analyzing GNSS observations as a spatio-temporal modeling problem. Compared to previous works, we construct a transformer-like attention mechanism to enhance the long short-term memory (LSTM) networks, improving model performance and generalization. For the training and evaluation of the proposed network, we used labeled datasets from the cities of Hong Kong and Aachen. We also introduce a dataset generation process to label the GNSS observations using lidar maps. In experimental studies, we compare the proposed network with a deep-learning-based model and classical machine-learning models. Furthermore, we conduct ablation studies of our network components and integrate the NLOS detection with data out-of-distribution in a state estimator. As a result, our network presents improved precision and recall ratios compared to other models. Additionally, we show that the proposed method avoids trajectory divergence in real-world vehicle localization by classifying and excluding NLOS observations.
\end{abstract}

\section{INTRODUCTION} \label{sec: intro}
On the way towards accurate and reliable localization of transport systems in outdoor and large-scale environments, global navigation satellite systems (GNSS) are widely used as standalone positioning solutions or sensor observations in advanced state estimation algorithms \cite{groves}. In the configuration of Real-Time Kinematic (RTK) with GNSS-correction data from a local ground-based augmentation service (GBAS), centimeter accuracy can be achieved \cite{gnss-rtk_analysis}. However, the performance of GNSS-based localization approaches is highly dependent on sufficient satellite visibility and low environmental interference. These requirements are frequently violated in complex environments such as urban canyons and bridge-rich areas \cite{integrity_survey}. In such environments, infrastructure objects can easily block GNSS measurements, resulting in non-line-of-sight (NLOS) receptions and multipath effects in the received GNSS observations \cite{multipath_NLOS_Groves}. Compared to atmospheric influences or satellite orbit bias, which can be corrected by the GNSS correction and ephemeris data, NLOS signals and multipath delays cannot be adequately modeled by statistical methods or eliminated using reference measurements from an external data link \cite{hsu_analysis_NLOS}. Therefore, consistent and robust vehicle localization in urban areas remains a challenge. 

Many earlier works rely predominantly on statistical error modeling or consistency checking within the obtained GNSS observations to exclude strongly corrupted GNSS measurements for vehicle positioning. In \cite{zhang_multipath_modeling} and \cite{comp_multipath_modeling}, multipath errors are indirectly modeled using signal strength measures such as signal-to-noise ratio ($\mathrm{SNR}$) and carrier-to-noise ratio ($\mathrm{C/N}_0$). However, such methods do not scale well in environments where multipath reflections with strong signal strength exist \cite{hsu_analysis_NLOS}. In contrast, the receiver autonomous integrity monitoring (RAIM) approach detects GNSS measurement faults by checking the consistency of received redundant GNSS observations. After computing a protection level that yields a pre-defined integrity risk and alarm limit, the RAIM indicates faulty measurements by utilizing online statistical tests \cite{wraim}. Although RAIM has become the prime technology in current GNSS solutions for fault isolation, this concept still assumes Gaussian noise distributions of GNSS observations. Moreover, it requires sufficient measurement redundancy, which cannot be fulfilled in urban areas and therefore leads to conservative fault classifications \cite{integrity_survey}. Other prominent works focus on state estimation algorithms that utilize robust error functions, such as m-estimators, or integrate other sensors to mitigate the effect of degraded GNSS measurements. In \cite{ekf_fgo_compare} and \cite{fgo_zhang}, the authors have proposed a tightly coupled fusion of inertial measurement alongside corrupted GNSS observations based on factor graph optimization with robust noise modeling using m-estimators and have shown superior performance against classic Bayesian filters. The principle of Graduated-Non-Convexity instead of traditional m-estimators can further mitigate the impact of strongly corrupted GNSS observations in urban areas \cite{fgo_gnc}. However, these approaches cannot explicitly exclude faulty GNSS measurement and thus do not yet guarantee stability and robustness.

Another group of approaches employs prior environmental information, such as 3D city models or fish-eye images, to acquire a satellite-blocking mask in the sky plot that is conducted for further visibility argumentation \cite{3dma_1, 3dma_2, 3dma_3, fisheye_1, fisheye_2}. Also, lidar sensing, which is usually used as an optical vehicle odometer, has been explored for GNSS fault exclusion. The generated 3D point cloud map can provide a more detailed and effective satellite-blocking mask considering local objects such as trees and large vehicles. This mask can be used for GNSS fault exclusion in state estimators, \cite{lidar1, lidar2}. However, these approaches rely on well-calibrated third-party sensors or prior information and cannot always guarantee real-time efficiency for fault detection. 

Recently, learning-based methods have drawn great attention for NLOS classification and multipath error prediction \cite{classical_learning}. Pre-trained models can be inferred online for GNSS fault detection after intensive data augmentation and offline training. Some works have utilized classical machine learning methods such as fuzzy logic \cite{classical_learning_0} or support vector machines \cite{classical_learning_1, classical_learning_2} to classify faulty GNSS measurements, which achieve impressive classification accuracy in test data. Considering the time-correlated and strongly nonlinear properties of faulty GNSS observations, deep-learning-based (DL) methods can extract semantic features and acquire better generalization outcomes using pre-trained models \cite{dl1, dl2, learning_survey}. Among all of these works, Zhang et al. proposed a particular network architecture containing fully-connected layers and long short-term memory (LSTM) networks for satellite visibility classification and pseudorange error prediction \cite{lstm_base}. In \cite{lstm_base_application}, the pre-trained network in \cite{lstm_base} has been utilized in a state estimator to exclude the NLOS observations and shows promising accuracy improvement. Although LSTM networks are considered a feasible method to model sequential data \cite{lstm_ts}. Therefore, they outperform other approaches for NLOS detection \cite{learning_survey}. However, LSTM networks are limited in applications with multivariate features and contextual information \cite{lstm_bad}. To overcome this problem, the self-attention mechanism introduced as transformers can be utilized to improve performance in sequential data modeling \cite{attention}. 

Inspired by \cite{lstm_base} and \cite{lstm_ts}, we consider the GNSS observations as time-series data and propose a transformer-enhanced LSTM network for NLOS classification and pseudorange error prediction. We follow the noise characteristics of the GNSS observations in complex environments and consider both temporal and spatial information in the feature extraction process. To train the network, we use the data set in \cite{lstm_base} and introduce a data generation process, which is used to label the GNSS data from our measurement campaign in the city of Aachen. We train the proposed network alongside a baseline LSTM network in \cite{lstm_base} and two classical machine learning models: support vector machine (SVM) and extreme gradient boosting (XGBoost). All models are evaluated with test and unseen (out-of-distribution) data. 
We also conduct ablation studies on our network components and present comprehensive discussions on model performance, data set quality, and the importance of features. Furthermore, we employ the pre-trained models on unlabeled data from our measurement campaign in the city of Düsseldorf to infer NLOS classifications, which are integrated into a state estimator. 

For the benefit of the research community, we release our code and Aachen dataset in \texttt{github}\footnotemark[1].

\footnotetext[1]{\url{https://github.com/rwth-irt/DeepNLOSDetection}}

\section{Dataset Generation} \label{sec:data}
\subsection{Data Collection}
For the training and testing of the learning-based algorithm, we collected raw data in a measurement campaign in Aachen and proposed a data labeling process. The measurement campaign contains a route with a length of $\SI{17}{\km}$ in different urban areas. The dataset consists of raw measurements from an inertial measurement unit (IMU) at $\SI{100}{\Hz}$, lidar point clouds at $\SI{10}{\Hz}$, and GNSS measurements at $\SI{10}{\Hz}$ using a NovAtel PwrPak7 receiver with RTCMv3 correction data received from the German GBAS server\footnotemark[2]. 
\footnotetext[2]{\url{https://sapos.de/}}

\subsection{Label Generation}
\begin{figure}[!ht]
    \centering
    \includegraphics[width=0.35\textwidth]{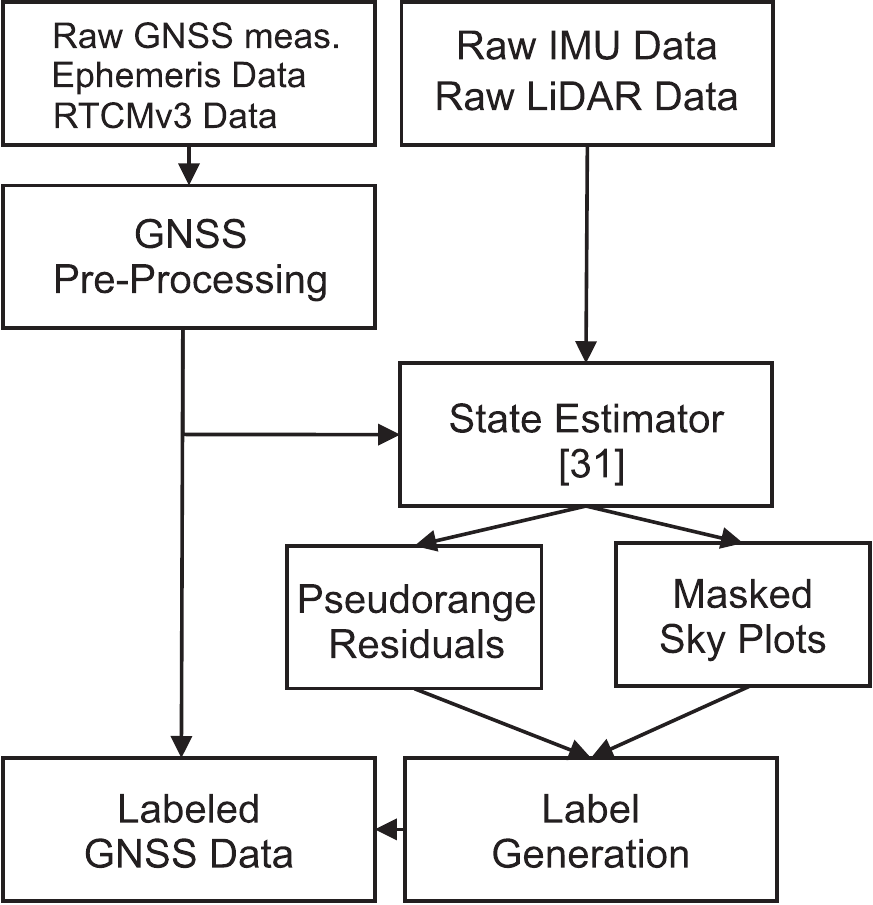}
    \caption{Proposed labeling framework}
    \label{fig:label_framework}
\end{figure}

\begin{figure}[!ht]
    \centering
    \includegraphics[width=0.48\textwidth]{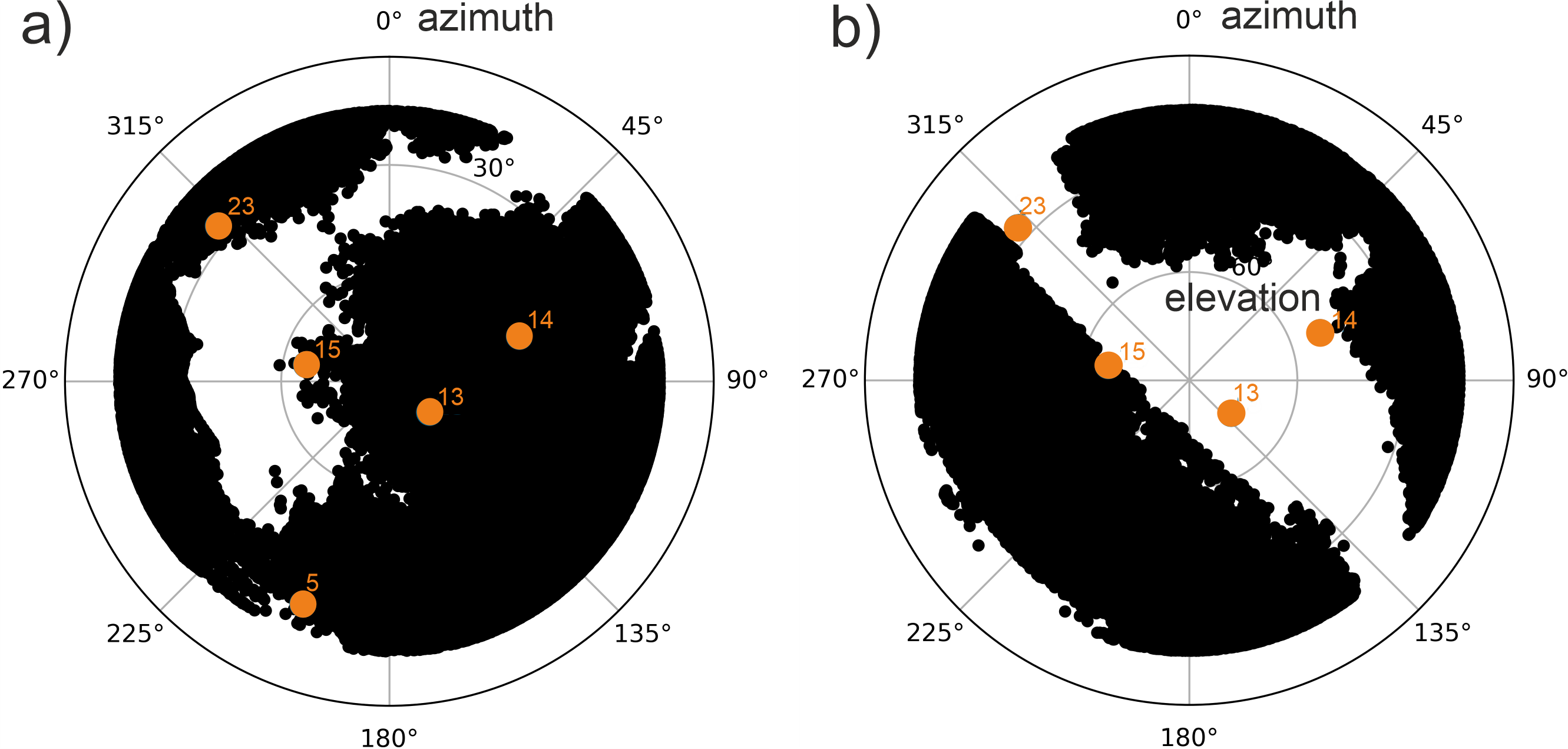}
    \caption{Sky plots from two scenes: a) urban tree-rich area, b) urban. The received GPS satellites are marked in orange.}
    \label{fig:sky_plots}
\end{figure}

Fig.\,\ref{fig:label_framework} shows our label generation process. To obtain the pseudorange residuals and satellite blocking masks, we use a multi-sensor fusion state estimator and a GNSS preprocessor from our previous work \cite{onlineFGO}. The raw GNSS measurements with the ephemeris data and RTCMv3 correction data are processed in the GNSS preprocessor to eliminate the satellite orbit bias and the atmospheric delays. After GNSS preprocessing, we feed the GNSS observations with the IMU measurements and lidar point clouds into the state estimator, where the pseudorange residuals are calculated and 3D lidar maps are generated. The 3D lidar maps are used to mark the satellite blockages in the sky plots, which are further conducted for label generation. Fig.\,\ref{fig:sky_plots} shows the masked sky plots. While generating labels, we query the neighbored elevation and azimuth angles at each satellite and infer the satellite with NLOS reception if the elevation/azimuth of the satellite is masked according to and the pseudorange residual is above a predefined threshold. 

\subsection{GNSS Feature Selection}
Following the previous work \cite{lstm_base}, we select the same five features from the pre-processed GNSS observations for the NLOS classification and pseudorange error prediction. 

\subsubsection{Elevation Angle $(\mathrm{El})$}
The satellite elevation angle $\mathrm{El}$ provides an adequate quality measure of the satellite visibility. Satellites with higher elevation angles are generally less affected by the infrastructure objects. In classical GNSS positioning approaches, satellites with low elevation angles are excluded from state estimation algorithms. The elevation angle $\mathrm{El}_k$ of $k$-th satellite can be calculated as
\begin{align}
    \mathrm{El}^k = \arcsin(\nicefrac{\myFrameScalar{h}{}{\mathrm{sat},k}}{r^k})\mathrm{,}
\end{align}
where the satellite's height with respect to the vehicle in the East-North-Up (ENU) frame is represented as $\myFrameScalar{h}{}{\mathrm{sat},k}$. The variable $r^k$ denotes the geometric range from the vehicle to the $k$-th satellite. 

\subsubsection{Azimuth Angle $(\mathrm{Az})$}
The satellite azimuth angle represents the horizontal geometric relation between the satellite and the vehicle. As described in \cite{lstm_base}, azimuth angles allow deducing the satellite visibility if multiple satellites with similar azimuth angles but different elevation angles are received. This feature is expected to provide more information on the geometric context in learning-based methods.
We calculate the azimuth angle with
\begin{align}
   \mathrm{Az}^k = \arctan(\nicefrac{x^{\mathrm{sat},k}}{y^{\mathrm{sat},k}}),
\end{align}
where the symbols $x^{\mathrm{sat},k}$ and $y^{\mathrm{sat},k}$ are the positions of $k$-th satellite relative to vehicle's position in the ENU-frame. 

\subsubsection{Carrier-to-Noise-Ratio $(\mathrm{C/N}_0)$}
The Carrier-to-Noise Ratio $\mathrm{C/N}_0$ measures the satellite signal strength and characterizes the satellite signal quality. Generally, the GNSS observations used for positioning are expected to have higher values $\mathrm{C/N}_0$. In urban scenarios, $\mathrm{C/N}_0$ can also effectively represent interference due to NLOS receptions and multipath effects \cite{hsu_analysis_NLOS}. 

\subsubsection{Least-Square Pseudorange Residual $(\sigma_{\mathrm{LS}}^{\mathrm{sat}})$}
In many GNSS standalone solutions, vehicle position can be calculated with an iterative least-square (LS) method using GNSS observations. We use this positioning solution to calculate pseudorange residuals, denoted as $\sigma_{\mathrm{LS}}^{\mathrm{sat},k}$. Unlike the pseudorange residuals calculated in the state estimator by combining multiple sensors, the residual $\sigma_{\mathrm{LS}}^{\mathrm{sat},k}$ is highly related to the quality of GNSS observations and represents the scale of pseudorange errors. We calculate the LS pseudorange residual by
\begin{align}
    \sigma_{\mathrm{LS}}^{\mathrm{sat},k} = \myFrameScalarHat{\rho}{k}{}-  \lVert \myFrameVecHat{x}{}{r} - \myFrameVec{x}{}{\mathrm{sat},k} \rVert_2,
\end{align}
where the range $\myFrameScalarHat{\rho}{k}{}$ is the measured pseudorange, which is corrected in the GNSS pre-processing process to eliminate the atmospheric delays. The satellite position and the estimated vehicle position are represented as $\myFrameVec{x}{}{\mathrm{sat},k}$ and $\myFrameVecHat{x}{}{r}$, respectively. 

\subsubsection{Root-Sum-Squares of Pseudorange Residuals $(\mathrm{RSS})$}
As introduced in \cite{lstm_base}, the root-sum-square of the pseudorange residuals encodes the environmental condition by summing up multiple pseudorange residuals in a certain time window. If the vehicle is moving in urban areas in $t=1...T$, larger $\mathrm{RSS}$ should be expected. We employ the $\mathrm{RSS}$, shown in
\begin{align}
    \mathrm{RSS} = \sqrt{\sum_{t=1}^T(\sigma_{\mathrm{LS},t}^{\mathrm{sat},k})^2}
\end{align}
as the last feature in the feature vector.

\section{Proposed Network}
\subsection{Problem Formulation}
The pseudorange observations are commonly used to represent the vehicle position in GNSS-based localization approaches, as shown in
\begin{align}
    \myFrameScalar{\rho}{t}{k} &= r_t^{\mathrm{sat},k} + c(\delta t_r - \delta t^{\mathrm{sat},k}) + T(t) + I(t) + M + \epsilon_k\mathrm{,} \label{eq: rho}\\
    \myFrameScalarHat{\rho}{t}{k} &= r_t^{\mathrm{sat},k} +  M + \epsilon_k \label{eq: df_rho}\mathrm{,}
\end{align}
where the range $r_t^{\mathrm{sat},k}=\lVert \myFrameVec{x}{t}{r} - \myFrameVec{x}{t}{\mathrm{sat},k} \rVert$ is the geometric range between the vehicle and $k$-th satellite. The receiver clock error $\delta t_r$ and the satellite clock error $\delta t^{\mathrm{sat},k}$, multiplied by the light speed $c$, represent the clock bias. The variables $T(t)$ and $I(t)$ are tropospheric and ionospheric delays, respectively. 

In this work, we assume that the receiver clock bias is estimated in the state estimator and therefore is subtracted from \eqref{eq: rho}. With the ephemeris and differential correction data, the satellite clock bias and atmospheric delays can be eliminated in a pre-processing process to acquire the corrected pseudorange $\myFrameScalarHat{\rho}{t}{k}$. We denote the multipath delay and the observation noise using $M$ and $\epsilon_k$, respectively. These delays vary particularly in urban areas and become complex and time-related due to NLOS receptions and multipath effects \cite{gnss_time_correlation}. 

Therefore, we consider the NLOS detection and error prediction of pseudorange $\myFrameScalarHat{\rho}{t}{k}$ as a spatio-temporal modeling problem
\begin{equation}
\label{eq: problem}
\begin{split}
    [\mathcal{V}^k,~\myFrameScalarHat{\epsilon}{}{k}] =  \mathrm{NN}\left(\left\{\mathrm{El}_{\tau}^k,~\mathrm{Az}_\tau^k,~\mathrm{C/N_0}_\tau^k,~\sigma_{\mathrm{LS},\tau}^k,~\mathrm{RSS}^k \right\}_{\tau = 1}^T\right)\mathrm{,}
\end{split}
\end{equation}
which predicts the satellite visibility $\mathcal{V}^k$ and pseudorange error $\myFrameScalarHat{\epsilon}{}{k}$ using a neural network given all selected features in a time series of $T$ seconds.     

\subsection{Neural Network Architecture}\label{sec: nn}
\begin{figure}[!ht]
    \centering
    \includegraphics[width=0.48\textwidth]{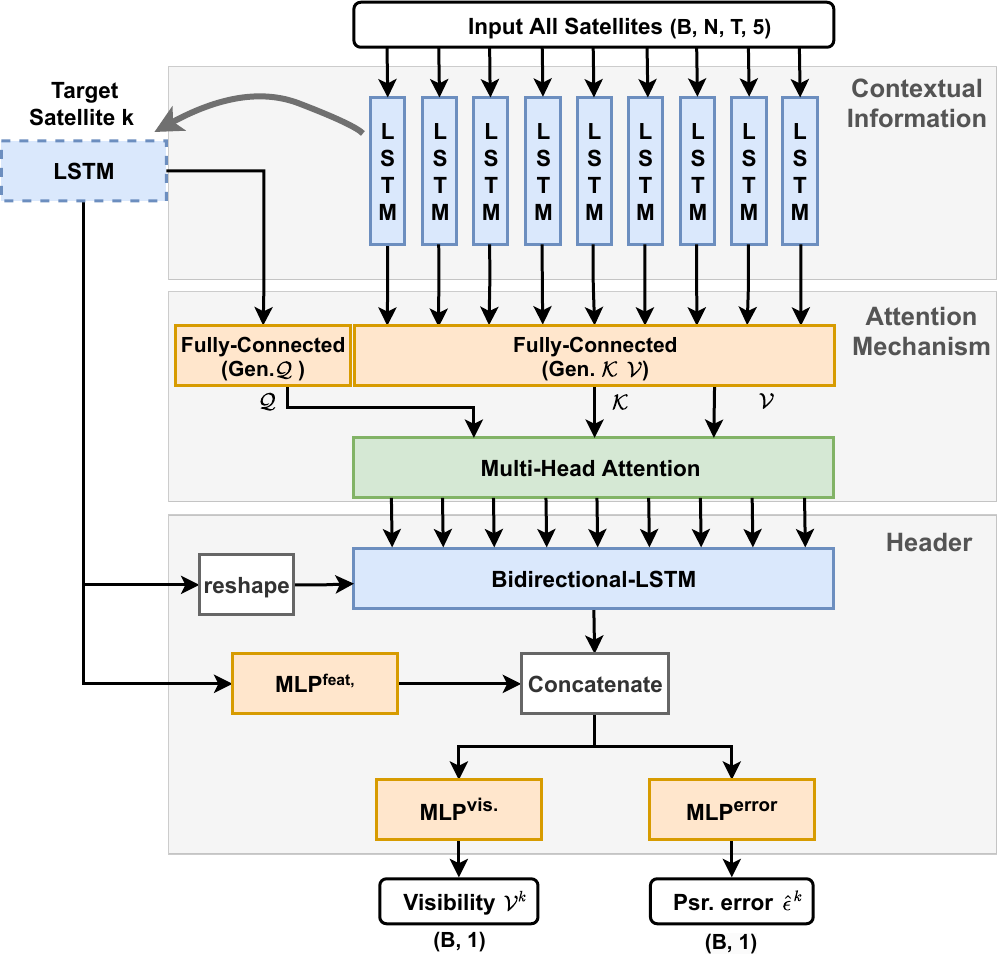}
    \caption{Proposed Neural Network Architecture.}
    \label{fig:nn}
\end{figure}
Considering the noise characteristics of GNSS observations in urban areas, we propose a neural network containing three properties: 1. capturing the temporal variations of selected features, 2. evaluating spatial contextual information from other satellites, and 3. providing effective attention for discriminative features. The proposed neural network architecture is presented in Fig.\,\ref{fig:nn}. 

Assuming that $N$ satellites can be received at maximum, we formulate the input tensor as a $T$ second time series of feature vectors from each satellite in a mini-batch $B$. To extract high-level features that contain temporal information from the received observations, we process the time series of each satellite with LSTM networks individually. The LSTM output features of all satellites are forwarded to a fully connected layer to generate contextual key $\mathcal{K}$ and value $\mathcal{V}$ features equivalent to transformer networks. Meanwhile, we detach the LSTM output features as input for three network components to obtain predictions for each target satellite. Above all, we employ the same fully-connected layer to generate the query features $\mathcal{Q}$ for the attention mechanism. The features $\mathcal{K},~\mathcal{V}$, and $\mathcal{Q}$ are evaluated in the multi-head attention sub-network, which is designed to propose spatial discriminative features from contextual information. In the header stack, we employ a bi-directional LSTM (Bi-LSTM) by initializing its hidden states with the output features from the foregoing LSTM network of the target satellite. The output features of the attention mechanism are processed in the Bi-LSTM, which is expected to extract the temporal information of the discriminative features and mitigate the effect of satellite ordering in the input tensor. We concatenate the output of the Bi-LSTM network with LSTM features of each target satellite to obtain high-level semantic features for classification and regression problems, where a binary visibility indicator $\mathcal{V}^k$ and a predicted pseudorange error $\myFrameScalarHat{\epsilon}{}{k}$ of the target satellite $k$ are inferred, as shown in Fig.,\ref{fig:nn}. 

Because our input tensor does not contain high-dimensional features, LSTM networks are constructed with two hidden layers. We implement the MLPs with three hidden layers and use $\mathrm{ReLU}$ as the activation function. For more details, see the code introduced in Sec.\,\ref{sec: intro}.

\subsection{Implementation and Training}
\subsubsection{Training and Test Data}
In this work, we use the Hong Kong (HK) dataset \cite{lstm_base} and our labeled dataset in Aachen (AC) for training and testing. All dataset details are shown in Tab.\,\ref{tab: data_details}. Compared to the HK dataset, where more NLOS observations can be received due to the high urbanization rate, the AC dataset only contains $\SI{18.87}{\percent}$ of NLOS receptions, presenting an unbalanced dataset. To evaluate the model generalization with data out-of-distribution, we use unlabeled data recorded in Düsseldorf (DUS) and predict the satellite visibility. This prediction is validated using a state estimation algorithm. 

\begin{table}[!t]
    \centering
    \caption{Details of Datasets. The average and maximal number of the received satellite are denoted with $n_{\mathrm{avg}}^{\mathrm{sat}}$ and $n_{\mathrm{max}}^{\mathrm{sat}}$. The maximal pseudorange error $\sigma_{\mathrm{max}}^{\rho}$ is inferred with ground-truth positioning. We present the ratio of LOS and NLOS labels in the dataset with $R_{}^{\mathrm{LOS}}$ and $R_{}^{\mathrm{NLOS}}$. }
    \resizebox{0.48\textwidth}{!}{\begin{tabular}{c|c|c|c|c|c|c}
    \hline
     Dataset &  \begin{tabular}[c]{@{}c@{}}Duration\\ \si{\second} \end{tabular}  & $n_{\mathrm{avg}}^{\mathrm{sat}}$ & $n_{\mathrm{max}}^{\mathrm{sat}}$ & \begin{tabular}[c]{@{}c@{}}$\sigma_{\mathrm{max}}^{\rho}$ \\ \si{\meter}\end{tabular} & \begin{tabular}[c]{@{}c@{}}$R_{}^{\mathrm{LOS}}$ \\ \si{\percent}\end{tabular} &  \begin{tabular}[c]{@{}c@{}}$R_{}^{\mathrm{NLOS}}$ \\ \si{\percent}\end{tabular}\\
      \hline
      HK \cite{lstm_base}  & 10573 & 17 & 25 & 1058.13 & 60.62 & 39.38\\
      \hline
      AC                   & 2366 & 6  & 8  & 640.32 & 81.13 & 18.87\\
      \hline
      DUS\footnotemark[3]  & 810  & 7  & 11 & 355.74 & - & - \\
      \hline
    \end{tabular}}
    \label{tab: data_details}
\end{table}
\footnotetext[3]{Only used for prediction.}

\subsubsection{Training Protocol}
We implement all models using \texttt{PyTorch}\footnotemark[4]. deep-learning (DL) models are trained with the \texttt{Adam} optimizer \cite{Kingma2014AdamAM} by utilizing a multistep learning rate scheduler. In the implementation, we use the $\mathrm{MSE}$ loss for the NLOS classification and $\mathrm{L1}$ loss in the regression problem to predict the pseudorange error. 

We employ the basic implementation in \texttt{Sklearn}\footnotemark[5] for all classical machine-learning models. The SVM classifier was trained using the $\mathrm{RBF}$ kernel with the penalty parameter $C=1.0$. We implement XGBoost with $200$ decision trees and a learning rate $\alpha=1.0$ for the classification task. 

The machine learning models are only trained for NLOS classification, whereas the DL models conduct two loss functions both for NLOS classification and pseudo-range error prediction. We train all models with datasets from Aachen and Hong Kong and consider testing procedures using data from the same and different datasets, with the aim of validating the generalization of the model.

\footnotetext[4]{\url{https://pytorch.org/}}
\footnotetext[5]{\url{https://scikit-learn.org/}}
\section{Experimental Results}
\subsection{General Performance Metrics for NLOS Classification}
\subsubsection{Data In-Distribution}
To evaluate the superiority of the proposed network, we also train different classical machine-learning approaches and the network proposed in \cite{lstm_base} in the experimental study. Tab.\,\ref{tab: metrics_hk_hk} and Tab.\,\ref{tab: metrics_ac_ac} show the general performance metrics of all models using the same datasets for training and testing. 

Because the HK dataset presents well-balanced NLOS observations compared to the AC dataset, as shown in Tab.\,\ref{tab: data_details}, all models can achieve fair performance for both LOS and NLOS classifications. Among all, the classical machine-learning models also present high precision and recall ratios and slightly outperform the model introduced in \cite{lstm_base}. Our implementation of the model \cite{lstm_base} can reproduce similar performance metrics and shows higher precision due to different training protocols. Compared to \cite{lstm_base}, our proposed model with the attention mechanism outperforms all other models and presents both high precision and recall ratios. 

We employ the same training and testing configurations using the AC dataset that contains major LOS observations. The performance metrics are shown in Tab.\,\ref{tab: metrics_ac_ac}. In contrast to the LOS observations, which can be classified with high accuracy and recall ratios, the performance for the NLOS classification of all models degrades dramatically. Classical models like SVM are more sensitive to imbalanced datasets than DL models. This has also been discussed in \cite{svm_imbanlence}. To validate this conclusion, we retrain the SVM using a modified AC dataset containing equally sampled NLOS and LOS data, denoted as SVM$^*$. It can be shown from this experiment that the SVM trained with a balanced dataset shows improved performance compared to other models. On the contrary, DL models present expected learning performance even when trained on unbalanced datasets.

\begin{table}[!t]
\centering
\caption{Performance Metrics using HK Dataset.}
\begin{tabular}{c c c c c c}
               \hline
               Model &&   Precision & Recall &F1-Score &Acc.\\
                \cline{1-6} 
                \multirow{2}{*}{SVM} &LOS &0.87&0.86&0.86 &\multirow{2}{*}{0.82}\\
                &NLOS &0.72&0.74&0.73\\
                \hline
                \multirow{2}{*}{XGboost} &LOS &0.90&0.88&0.89 &\multirow{2}{*}{0.85}\\
                &NLOS &0.77&0.80&0.78\\
                \hline
                \multirow{2}{*}{DL \cite{lstm_base}} &LOS &0.80&0.96&0.87 &\multirow{2}{*}{0.81}\\
                &NLOS &0.85&0.51&0.64\\
                \hline
                \multirow{2}{*}{DL (ours)} &LOS &0.90&0.91&0.90 &\multirow{2}{*}{0.87}\\
                &NLOS &0.80&0.79&0.80\\
                \hline
            
 \end{tabular}
\label{tab: metrics_hk_hk}
\end{table}

\begin{table}[!t]
\centering
\caption{Performance Metrics using AC Dataset.}
\begin{tabular}{c c c c c c}
               \hline
               Model &&  Precision & Recall &F1-Score &Acc.\\
                \cline{1-6}   
                \multirow{2}{*}{SVM} &LOS &0.80&1.00&0.89 &\multirow{2}{*}{0.80}\\
                &NLOS &0.39&0.01&0.01\\
                \hline
                \multirow{2}{*}{SVM$^{*}$} &LOS &0.91&0.72&0.80 &\multirow{2}{*}{0.72}\\
                &NLOS &0.38&0.70&0.49\\
                \hline
                \multirow{2}{*}{XGboost} &LOS &0.86&0.88&0.87 &\multirow{2}{*}{0.79}\\
                &NLOS &0.47&0.42&0.45\\
                \hline
                \multirow{2}{*}{DL \cite{lstm_base}} &LOS &0.86&0.83&0.84 &\multirow{2}{*}{0.76}\\
                &NLOS &0.40&0.47&0.43\\    
                \hline
                \multirow{2}{*}{DL (ours)} &LOS &0.86&0.90&0.88 &\multirow{2}{*}{0.81}\\
                &NLOS &0.51&0.42&0.46\\
                \hline
            
 \end{tabular}
\label{tab: metrics_ac_ac}
\end{table}

\subsubsection{Data Out-Of-Distribution}
Furthermore, we conducted the experiment to evaluate the performance of pre-trained models on data out-of-distribution. As the HK dataset contains a fair distribution of the NLOS and LOS observations, we train all models using the HK dataset and test them with the AC dataset. The results are shown in Tab.\,\ref{tab: metrics_hk_ac}. In this experiment, the SVM and the proposed DL slightly outperform other models and show similar performance compared to each other. However, all models cannot show fair performance in the NLOS classification compared to Tab.\,\ref{tab: metrics_ac_ac}, where the classification precision and especially the recall ratio are more elevated. 

\begin{table}[!t]
\centering
\caption{Cross-Dataset Performance Metrics: The models are trained using the HK dataset and tested with the AC dataset.}
\begin{tabular}{c c c c c c}
               \hline
               Model & &  Precision & Recall &F1-Score &Acc.\\
                \cline{1-6}   
                \multirow{2}{*}{SVM} &LOS &0.88&0.82&0.85 &\multirow{2}{*}{0.77}\\
                &NLOS &0.42&0.54&0.47\\
                \hline
                \multirow{2}{*}{XGboost} &LOS &0.87&0.74&0.80 &\multirow{2}{*}{0.70}\\
                &NLOS &0.32&0.54&0.40\\
                \hline
                \multirow{2}{*}{DL \cite{lstm_base}} &LOS &0.86&0.81&0.84 &\multirow{2}{*}{0.74}\\
                &NLOS &0.35&0.45&0.39\\
                \hline
                \multirow{2}{*}{DL (ours)} &LOS &0.89&0.79&0.84 &\multirow{2}{*}{0.75}\\
                &NLOS &0.40&0.60&0.48\\
                \hline
            
 \end{tabular}
\label{tab: metrics_hk_ac}
\end{table}

\textbf{\textit{Discussion:}} For the NLOS classification task, we show that all learning-based models perform well by training with a well-balanced dataset. Even classical models present both high-precision and recall ratios. One possible reason for this result is that these models fit overwhelmingly on some features, which are highly related to NLOS receptions. In contrast, the contextual features that indirectly infer GNSS faults are discounted. If the classical models are inferred with data out-of-distribution, the recall ratio drops dramatically. Thus, we expect that the DL networks can present more generalized models. When comparing the DL models, our network is optimized with an attention mechanism that effectively captures the spatial and temporal information from the GNSS observations, showing better performance. However, the proposed method presents higher model complexity, which requires a more thorough training procedure and penalizes runtime efficiency. Considering the dataset quality, we emphasize the importance of label balance in the datasets, which plays a vital role in the performance of learning-based methods. We also show the performance degradation of all models while testing using data out-of-distribution. One probable reason can be the inherent quantity difference between the two datasets. For instance, we calculate the maximum, minimum, and average $\mathrm{C/N_0}$ in two datasets used for training, shown in Tab.\,\ref{tab: cn0_analysis}. Suppose that the model has been pre-trained using a dataset with data bias. In that case, the model will present a distribution shift, leading to performance degradation, as discussed in \cite{dist_shift}.

\begin{table}[!h]
    \centering
    \caption{Analysis of $\mathrm{C/N_0}$ in Two Datasets.}
    {\begin{tabular}{c|c|c|c}
    \hline
     Dataset &  \begin{tabular}[c]{@{}c@{}}max. $\mathrm{C/N_0}$ \\ \si{\dB}-\si{\hertz} \end{tabular} & \begin{tabular}[c]{@{}c@{}}min. $\mathrm{C/N_0}$ \\ \si{\dB}-\si{\hertz} \end{tabular} & \begin{tabular}[c]{@{}c@{}}avg. $\mathrm{C/N_0}$ \\ \si{\dB}-\si{\hertz} \end{tabular}\\
      \hline
      HK \cite{lstm_base}  & 49.0 & 7.0 & 34.0\\
      \hline
      AC                   & 57.0 & 22.0  & 49.0 \\
      \hline
    \end{tabular}}
    \label{tab: cn0_analysis}
\end{table}
\begin{figure}[!t]
    \centering
    \includegraphics[width=0.4\textwidth]{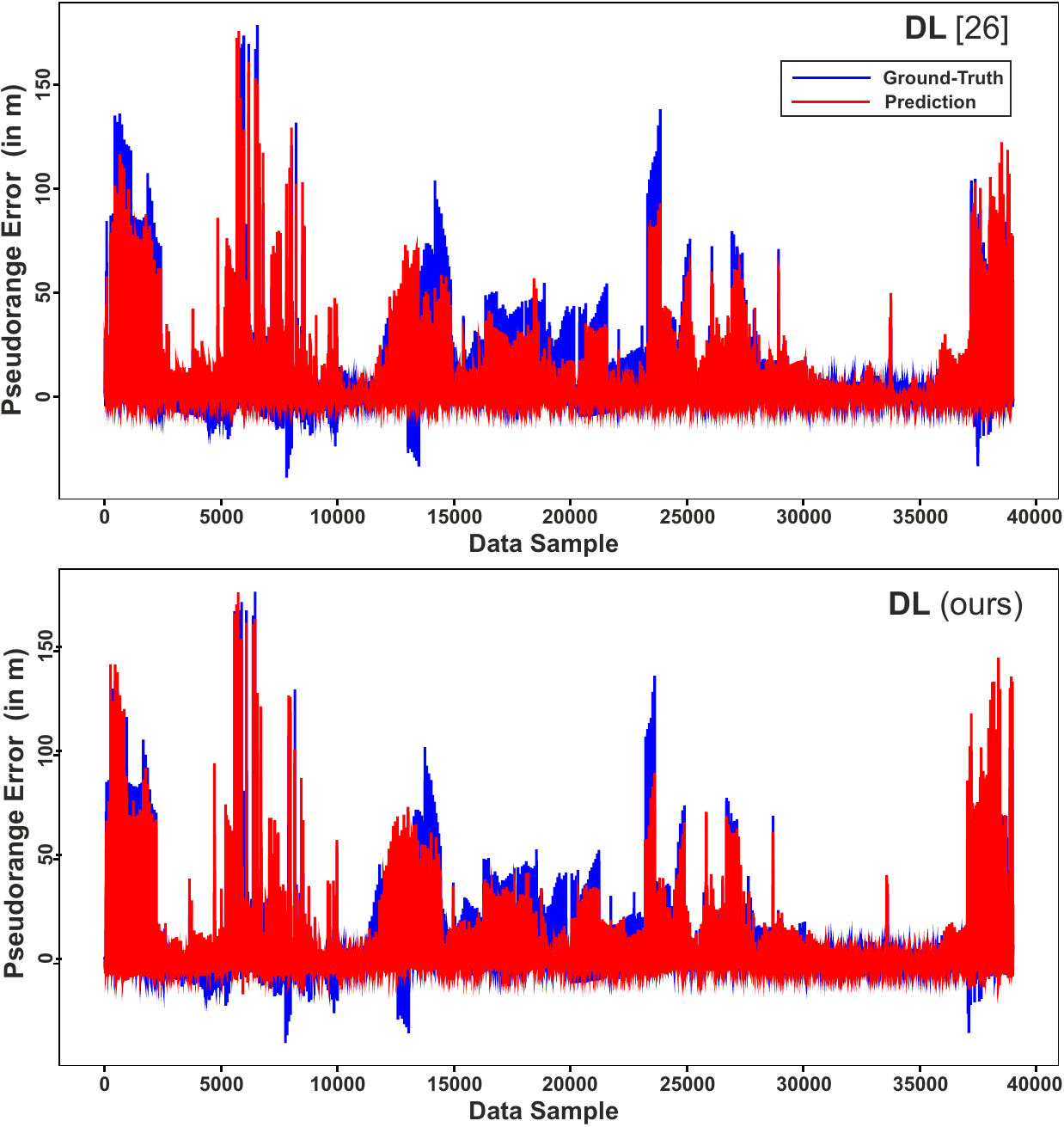}
    \caption{Predicted and ground-truth pseudorange errors with the model \cite{lstm_base} and our network.}
    \label{fig: psr_errors}
\end{figure}

\subsection{Performance of Pseudorange Error Prediction}
We only evaluated the baseline DL models in the pseudorange error prediction task. The predicted pseudorange errors and ground-truth are illustrated in Fig.\,\ref{fig: psr_errors}, presenting a qualitative performance view. Compared to positive pseudorange errors, both models do not perform well in predicting negative and large pseudorange errors, which have a non-negligible impact on consistent state estimation. The average prediction error when testing the proposed model amounts to $\SI{5.68}{\meter}$, which is lower than $\SI{6.53}{\meter}$ from the model in \cite{lstm_base}. 

\textbf{\textit{Discussion:}} Even though the proposed network presents a higher prediction accuracy, we indicate that the current pre-trained DL models are not able to infer plausible pseudorange error predictions. Because there are multiple interference sources for the pseudorange, faulty pseudoranges present highly complex noise distributions. Thus, the most probable reason is the insufficient data samples of faulty pseudoranges with large errors. This could be improved if the network structure for the regression task is further optimized and more faulty GNSS observations can be sampled for training.

\subsection{Ablation Studies}
Besides general performance metrics, we propose ablation studies on feature importance and different components of our DL model. 

\subsubsection{Feature Importance} \label{sec: ab_feature}
\begin{figure}[!t]
    \centering
    \includegraphics[width=0.48\textwidth]{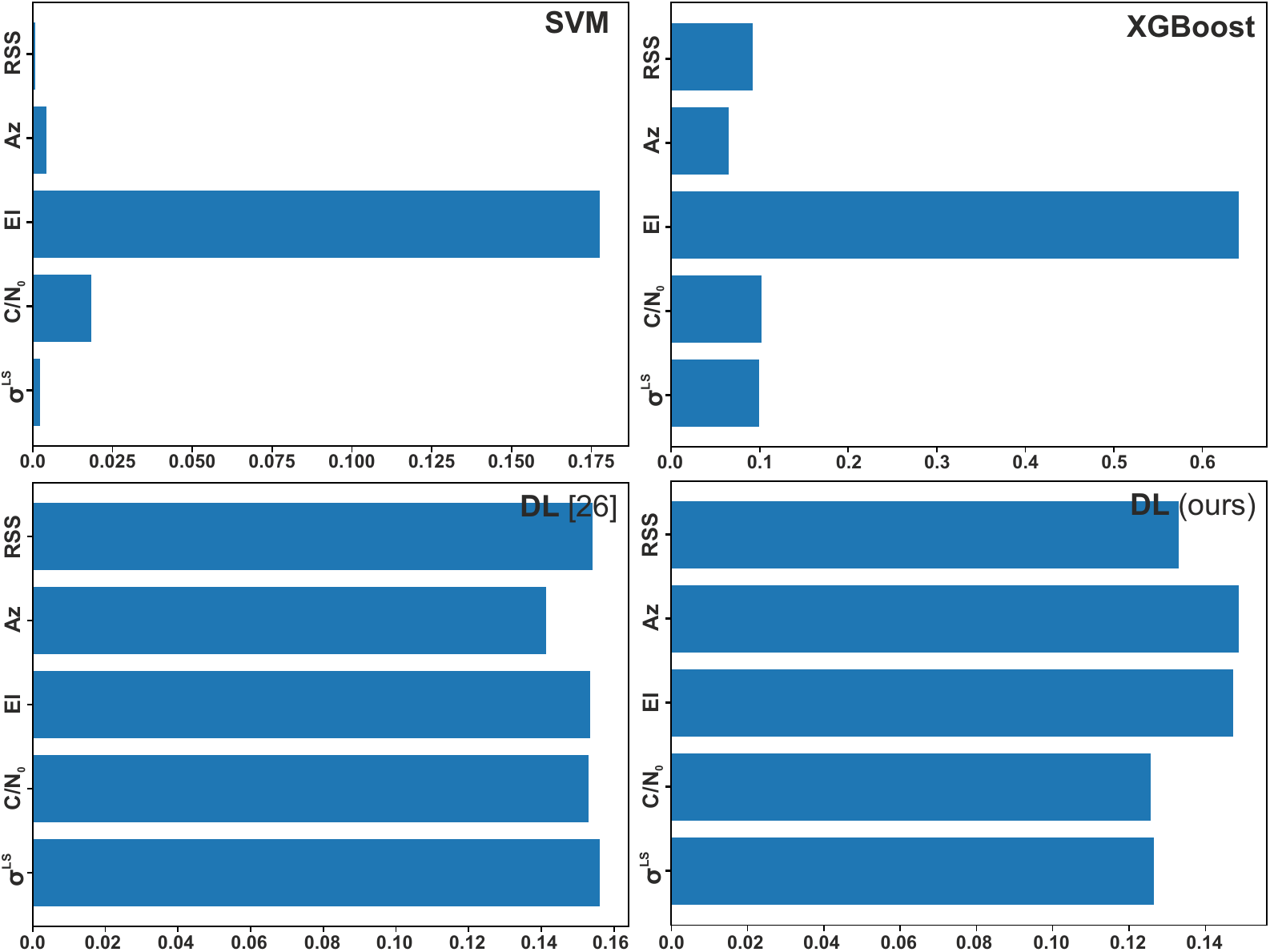}
    \caption{Feature Permutation Importance of different models.}
    \label{fig: feature_importance}
\end{figure}

Although the classical models can achieve the same high accuracy as the DL models when comparing the results in Tab.\,\ref{tab: metrics_hk_hk} and Tab.\,\ref{tab: metrics_ac_ac}, the DL models show better generalization considering unbalanced or unseen data. One reason to support this hypothesis can be speculated with the feature permutation importance, as shown in Fig.\,\ref{fig: feature_importance}. While classical models significantly rely on a few features, such as elevation angle and $\mathrm{C/N}_0$, DL models tend to rate all features equally. This behavior should be expected in this learning problem because features such as azimuth angle and the Root-Sum-Squares (RSS) provide crucial contextual information about signal blockage in urban areas. A fair weighting of all features can contribute more to model generalization and, thus, avoid the overfitting effect.

\subsubsection{Model Components} \label{sec: ab_components}
To prove our hypothesis in the network design, we evaluated different components of our model and discuss the effect of integrating the attention mechanism and the Bi-LSTM network. For this experiment, the HK dataset and identical training settings as in Tab.\,\ref{tab: metrics_hk_hk} were employed to propose the results in Tab.\,\ref{tab: ablation_study}. 

As introduced in Sec.\,\ref{sec: nn}, handling spatial information using the attention mechanism benefits both model precision and recall ratio for the NLOS classification, even when the dataset contains more LOS observations. Replacing the Bi-LSTM with a normal LSTM network penalizes the recall ratio for NLOS detection. This phenomenon confirms the superiority of the Bi-LSTM, as introduced in \cite{bi_lstm}, that the Bi-LSTM shows a large model capacity by preserving information from both the past and future, especially when contextual information is non-negligible.
\begin{table}[!t]
\caption{Performance metrics during ablation studies of the proposed method.}
\begin{tabular}{c c c c c c c}
               \hline
               Model & &   Precision & Recall &F1-Score &Acc.\\
                \cline{1-6}   
                \multirow{2}{*}{w/o Att. Mecha.} &LOS &0.83&0.91&0.87 &\multirow{2}{*}{0.82}\\
                &NLOS &0.76&0.63&0.69\\
                \hline
                \multirow{2}{*}{w/o BiLSTM} &LOS &0.81&0.91&0.86 &\multirow{2}{*}{0.80}\\
                &NLOS &0.76&0.57&0.65 \\
                \hline
            
 \end{tabular}
\label{tab: ablation_study}
\end{table}

\subsection{Vehicle Localization with GNSS Fault Exclusion}
\begin{figure*}[!ht]
    \centering
    \includegraphics[width=1\textwidth]{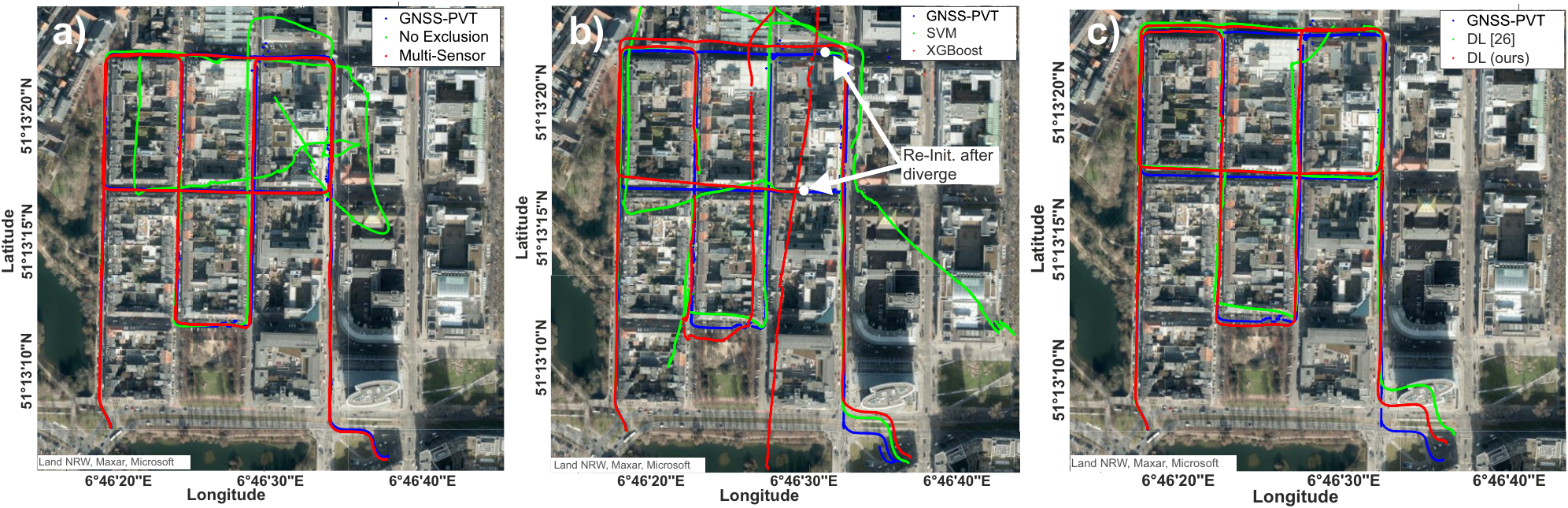}
    \caption{Vehicle Localization in Urban Area in Düsseldorf with NLOS Exclusion. The GNSS Positioning-Velocity-Timing (PVT) solution is shown in blue. a) presents the estimated trajectory without NLOS exclusion and the solution by fusing GNSS observations and lidar odometry in a tight coupling. b) and c) illustrate the estimated trajectories with NLOS exclusion using different learning models.}
    \label{fig: trajectory_dus}
\end{figure*}
In this experiment, we employ the pre-trained models to infer the NLOS receptions of GPS-L1 and Galileo-E1 observations obtained using a NovAtel PwrPak7 receiver in a measurement campaign in Düsseldorf. The GNSS observations are not used for training and, thus, present data out-of-distribution. By employing the pre-trained model using HK and AC datasets, the predicted ratio of NLOS observations of each model is shown in Tab.,\ref{tab: prediction_dus}. 

\begin{table}[!h]
    \centering
    \caption{Result of NLOS classification with data out-of-distribution.}
    \begin{tabular}{c|c|c}
    \hline
     Model &  \begin{tabular}[c]{@{}c@{}}$\hat{R}_{}^{\mathrm{LOS}}$ \\ \si{\percent}\end{tabular} &  \begin{tabular}[c]{@{}c@{}}$\hat{R}_{}^{\mathrm{NLOS}}$ \\ \si{\percent}\end{tabular}\\
      \hline
      SVM       & 64.83 & 35.17\\
      \hline
      XGBoost    & 58.68 & 41.32\\
      \hline
      DL \cite{lstm_base}                 & 76.74 & 23.26 \\
      \hline
      DL (ours)                & 74.76 & 25.24\\
      \hline
    \end{tabular}
    \label{tab: prediction_dus}
\end{table}

We integrate the NLOS detections of all models in a graph-optimization-based state estimation algorithm that fuses the GNSS pseudorange and Doppler shift with IMU measurements in a tight coupling \cite{onlineFGO}. Here, we consider the dual-constellation of GPS and Galileo measurements, which are preprocessed with RTCM correction data. The pseudorange observations, which are classified as NLOS receptions, are excluded for integration into the graph.

As shown in Fig.\,\ref{fig: trajectory_dus} a), the GNSS observations in challenging environments are strongly corrupted. Thus, the state estimator integrates observations with inconsistent sensor measurement models, leading to a divergence in trajectory estimation. This problem can be alleviated if the NLOS observations are effectively excluded, illustrated in Fig.\,\ref{fig: trajectory_dus} c).

Compared to DL models, more NLOS observations are identified using SVM and XGBoost. This reduces constraints that are factorized from the GNSS observations in the optimization problem, leading to dramatic trajectory drifting and divergence, as shown in Fig.,\ref{fig: trajectory_dus} b). Our hypothesis of this problem can be referred to Sec.\,\ref{sec: ab_feature}. The classical models generally count on a few features. They are not capable of extracting and evaluating high-level features, resulting in overconfident feature weighting, and thus, proposing more False-Positive NLOS classifications.  The DL models identify more LOS observations, which contributes to more sensor observations integrated for state estimation. On this basis, consistent trajectories are estimated if NLOS observations can be effectively classified, as demonstrated in Fig.,\ref{fig: trajectory_dus} c). 

\textbf{\textit{Discussion:}} In this experiment, our network infers more NLOS receptions compared to the model in \cite{lstm_base}. The vehicle trajectory can be consistently estimated by excluding the NLOS observations classified with our network, which indirectly confirms that the proposed network can infer more true-positive NLOS detections. However, the NLOS exclusion can improve the consistency of the trajectory estimation, but accuracy cannot yet be guaranteed. This result can be associated with the insufficient number of GNSS observations that survive after the NLOS exclusion in our measurement campaign. Thus, since the NLOS detector contributes a majorly to trajectory consistency, integrating other sensors into the state estimator still presents a feasible way to improve the precision of estimated trajectories in challenging environments.

\section{Conclusion and future work}
In this work, we proposed a transformer-enhanced LSTM network to detect NLOS receptions and predict pseudorange errors of GNSS observations. The proposed network was evaluated with a baseline LSTM network and classical machine learning models using two datasets from Hong Kong and Aachen. Besides, we introduced a dataset generation process using a multi-sensor state estimator. We conducted experimental studies of the pre-trained models both for data in-distribution and out-of-distribution. The results show that the deep-learning-based methods proposed more generalized models, which present considerable inference performance for the NLOS classification. Moreover, the DL models are less prone to overfitting compared to classical machine learning models. Among all models, the proposed network, enhanced by the attention mechanism, captures both temporal and spatial information with all satellite observations as contextual information and has achieved the best performance even with data out-of-distribution. Furthermore, our model shows effective NLOS detection in real-world vehicle localization, where the consistent trajectory can be estimated. For future work, more balanced datasets can be utilized to train the proposed network. And the input tensor can be extended with other advanced features, such as the Doppler rate consistency. We also plan to integrate the proposed method into our online state estimator for effective NLOS rejection \cite{onlineFGO}. 

\section*{Acknowledgments}
We sincerely thank Dr. Li-Ta Hsu, Dr. Guohao Zhang, Dr. Weisong Wen, and Penghui Xu from the Department of Aeronautical and Aviation Engineering at Hong Kong Polytechnic University for their generous data exchange and scientific discussions. We also thank Robin Taborsky from the Institute of Automatic Control at the RWTH Aachen University and the public order offices in Aachen and Düsseldorf for their support in the measurement campaigns.

\bibliographystyle{IEEETran.bst}
\bibliography{IEEEabrv, reference}

\begin{thebibliography}{10}
\providecommand{\url}[1]{#1}
\csname url@rmstyle\endcsname
\providecommand{\newblock}{\relax}
\providecommand{\bibinfo}[2]{#2}
\providecommand\BIBentrySTDinterwordspacing{\spaceskip=0pt\relax}
\providecommand\BIBentryALTinterwordstretchfactor{4}
\providecommand\BIBentryALTinterwordspacing{\spaceskip=\fontdimen2\font plus
\BIBentryALTinterwordstretchfactor\fontdimen3\font minus
  \fontdimen4\font\relax}
\providecommand\BIBforeignlanguage[2]{{%
\expandafter\ifx\csname l@#1\endcsname\relax
\typeout{** WARNING: IEEEtran.bst: No hyphenation pattern has been}%
\typeout{** loaded for the language `#1'. Using the pattern for}%
\typeout{** the default language instead.}%
\else
\language=\csname l@#1\endcsname
\fi
#2}}

\bibitem{groves}
P.~Groves, \emph{Principles of GNSS, Inertial, and Multisensor Integrated
  Navigation Systems, Second Edition}.\hskip 1em plus 0.5em minus 0.4em\relax
  Artech House, 2013.

\bibitem{gnss-rtk_analysis}
Y.~Feng and J.~Wang, ``{GPS} {RTK} performance characteristics and analysis,''
  \emph{Journal of Global Positioning Systems}, vol.~7, 06 2008.

\bibitem{integrity_survey}
N.~Zhu, J.~Marais, D.~Bétaille, and M.~Berbineau, ``{GNSS} position integrity
  in urban environments: A review of literature,'' \emph{IEEE Transactions on
  Intelligent Transportation Systems}, vol.~19, no.~9, pp. 2762--2778, 2018.

\bibitem{multipath_NLOS_Groves}
P.~Groves, ``{GNSS} solutions: Multipath vs. {NLOS} signals. how does
  {Non-Line-of-Sight} reception differ from multipath interference,''
  \emph{Inside GNSS (Magazine)}, vol.~8, pp. 40--42, 12 2013.

\bibitem{hsu_analysis_NLOS}
L.-T. Hsu, ``Analysis and modeling {GPS} {NLOS} effect in highly urbanized
  area,'' \emph{GPS Solutions}, vol.~22, 11 2017.

\bibitem{zhang_multipath_modeling}
Z.~Zhang, B.~Li, Y.~Gao, and Y.~Shen, ``Real-time carrier phase multipath
  detection based on dual-frequency {C/N0} data,'' \emph{GPS Solutions},
  vol.~23, 11 2018.

\bibitem{comp_multipath_modeling}
C.~Comp and P.~Axelrad, ``Adaptive {SNR}-based carrier phase multipath
  mitigation technique,'' \emph{IEEE Transactions on Aerospace and Electronic
  Systems}, vol.~34, no.~1, pp. 264--276, 1998.

\bibitem{wraim}
T.~Walter and P.~Enge, ``Weighted {RAIM} for precision approach,'' in
  \emph{Proc. 8th International Technical Meeting of the Satellite Division of
  The Institute of Navigation (ION GPS 1995)}, Sept. 24, 1995.

\bibitem{ekf_fgo_compare}
W.~Wen, T.~Pfeifer, X.~Bai, and L.-T. Hsu, ``Factor graph optimization for
  {GNSS/INS} integration: A comparison with the extended {Kalman} filter,''
  \emph{NAVIGATION}, vol.~68, no.~2, pp. 315--331, 2021.

\bibitem{fgo_zhang}
H.~Zhang, X.~Xia, M.~Nitsch, and D.~Abel, ``Continuous-{T}ime factor graph
  optimization for trajectory smoothness of {GNSS/INS} navigation in
  temporarily gnss-denied environments,'' \emph{IEEE Robotics and Automation
  Letters}, vol.~7, no.~4, pp. 9115--9122, 2022.

\bibitem{fgo_gnc}
W.~Wen, G.~Zhang, and L.-T. Hsu, ``{GNSS} outlier mitigation via graduated
  non-convexity factor graph optimization,'' \emph{IEEE Transactions on
  Vehicular Technology}, vol.~71, no.~1, pp. 297--310, 2022.

\bibitem{3dma_1}
S.~Xin, J.~Geng, G.~Zhang, H.-F. Ng, J.~Guo, and L.-T. Hsu,
  ``{3D}-mapping-aided {PPP-RTK} aiming at deep urban canyons,'' \emph{Journal
  of Geodesy}, vol.~96, 2022.

\bibitem{3dma_2}
H.-F. Ng, G.~Zhang, Y.~Luo, and L.-T. Hsu, ``Urban positioning: {3D}
  mapping-aided {GNSS} using dual-frequency pseudorange measurements from
  smartphones,'' \emph{NAVIGATION}, vol.~68, no.~4, pp. 727--749.

\bibitem{3dma_3}
H.-F. Ng, L.-T. Hsu, M.~J.~L. Lee, J.~Feng, T.~Naeimi, M.~Beheshti, and J.-R.
  Rizzo, ``Real-time loosely coupled {3DMA} {GNSS/Doppler} measurements
  integration using a graph optimization and its performance assessments in
  urban canyons of new york,'' \emph{Sensors}, vol.~22, no.~17, 2022.

\bibitem{fisheye_1}
X.~Bai, W.~Wen, and L.-T. Hsu, ``Using sky-pointing fish-eye camera and {LiDAR}
  to aid {GNSS} single-point positioning in urban canyons,'' \emph{IET
  Intelligent Transport Systems}, vol.~14, no.~8, pp. 908--914, 2020.

\bibitem{fisheye_2}
M.~J.~L. Lee, S.~Lee, H.-F. Ng, and L.-T. Hsu, ``Skymask matching aided
  positioning using sky-pointing fisheye camera and {3D} city models in urban
  canyons,'' \emph{Sensors}, vol.~20, no.~17, 2020.

\bibitem{lidar1}
\BIBentryALTinterwordspacing
X.~Liu, W.~Wen, F.~Huang, H.~Gao, Y.~Wang, and L.-T. Hsu, ``{3D} {LiDAR} aided
  {GNSS} {NLOS} mitigation for reliable {GNSS-RTK} positioning in urban
  canyons,'' 2022. [Online]. Available: \url{https://arxiv.org/abs/2212.05477}
\BIBentrySTDinterwordspacing

\bibitem{lidar2}
X.~Liu, W.~Wen, and L.-T. Hsu, ``{3D} {LiDAR} aided {GNSS} real-time kinematic
  positioning via coarse-to-fine batch optimization for high accuracy mapping
  in dense urban canyons,'' \emph{Proc. of the 35th International Technical
  Meeting of the Satellite Division of The Institute of Navigation (ION GNSS+
  2022)}, 2022.

\bibitem{classical_learning}
L.-T. Hsu, ``What are the roles of artificial intelligence and machine learning
  in {GNSS} positioning?'' \emph{Inside GNSS}, pp. 20--27, 2020.

\bibitem{classical_learning_0}
R.~Sun, L.-T. Hsu, D.~Xue, G.~Zhang, and W.~Y. Ochieng, ``{GPS} signal
  reception classification using adaptive neuro-fuzzy inference system,''
  \emph{The Journal of Navigation}, vol.~72, no.~3, p. 685–701, 2019.

\bibitem{classical_learning_1}
L.-T. Hsu, ``{GNSS} multipath detection using a machine learning approach,'' in
  \emph{2017 IEEE 20th International Conference on Intelligent Transportation
  Systems (ITSC)}, 2017, pp. 1--6.

\bibitem{classical_learning_2}
T.~Suzuki and Y.~Amano, ``{NLOS} multipath classification of {GNSS} signal
  correlation output using machine learning,'' \emph{Sensors}, vol.~21, no.~7,
  2021.

\bibitem{dl1}
Y.~Quan, L.~Lau, G.~W. Roberts, X.~Meng, and C.~Zhang, ``Convolutional neural
  network based multipath detection method for static and kinematic gps high
  precision positioning,'' \emph{Remote Sensing}, vol.~10, no.~12, 2018.

\bibitem{dl2}
R.~Zawislak, M.~Greiff, K.~Kim, K.~Berntorp, S.~Di~Cairano, M.~Konishi,
  K.~Parsons, P.~V.~Orlik, and Y.~Sato, ``{GNSS} multipath detection aided by
  unsupervised domain adaptation,'' \emph{Proc. of the 35th International
  Technical Meeting of the Satellite Division of The Institute of Navigation
  (ION GNSS+ 2022)}, pp. 2127--2137, 2022.

\bibitem{learning_survey}
A.~Siemuri, K.~Selvan, H.~Kuusniemi, P.~Valisuo, and M.~S. Elmusrati, ``A
  systematic review of machine learning techniques for {GNSS} use cases,''
  \emph{IEEE Transactions on Aerospace and Electronic Systems}, vol.~58, no.~6,
  pp. 5043--5077, 2022.

\bibitem{lstm_base}
G.~Zhang, P.~Xu, H.~Xu, and L.-T. Hsu, ``Prediction on the urban {GNSS}
  measurement uncertainty based on deep learning networks with long short-term
  memory,'' \emph{IEEE Sensors Journal}, vol.~21, no.~18, pp. 20\,563--20\,577,
  2021.

\bibitem{lstm_base_application}
X.~Bai, W.~Wen, G.~Zhang, H.-F. Ng, and L.-T. Hsu, ``{GNSS} outliers mitigation
  in urban areas using sparse estimation based on factor graph optimization,''
  in \emph{2022 IEEE 25th International Conference on Intelligent
  Transportation Systems (ITSC)}, 2022, pp. 197--202.

\bibitem{lstm_ts}
B.~Lim, S.~O. Arık, N.~Loeff, and T.~Pfister, ``Temporal fusion transformers
  for interpretable multi-horizon time series forecasting,''
  \emph{International Journal of Forecasting}, vol.~37, no.~4, pp. 1748--1764,
  2021.

\bibitem{lstm_bad}
Y.~Li, Z.~Zhu, D.~Kong, H.~Han, and Y.~Zhao, ``{EA-LSTM}: Evolutionary
  attention-based lstm for time series prediction,'' \emph{Knowledge-Based
  Systems}, vol. 181, p. 104785, 2019.

\bibitem{attention}
A.~Vaswani, N.~Shazeer, N.~Parmar, J.~Uszkoreit, L.~Jones, A.~N. Gomez, L.~u.
  Kaiser, and I.~Polosukhin, ``Attention is all you need,'' in \emph{Advances
  in Neural Information Processing Systems}, I.~Guyon, U.~V. Luxburg,
  S.~Bengio, H.~Wallach, R.~Fergus, S.~Vishwanathan, and R.~Garnett, Eds.,
  vol.~30.\hskip 1em plus 0.5em minus 0.4em\relax Curran Associates, Inc.,
  2017.

\bibitem{onlineFGO}
\BIBentryALTinterwordspacing
H.~Zhang, C.-C. Chen, H.~Vallery, and T.~D. Barfoot, ``{GNSS}/multi-sensor
  fusion using continuous-time factor graph optimization for robust
  localization,'' 2023. [Online]. Available:
  \url{https://arxiv.org/abs/2309.11134}
\BIBentrySTDinterwordspacing

\bibitem{gnss_time_correlation}
M.~Goode, S.~Edwards, and P.~Moore, ``Time correlation in {GNSS} precise point
  positioning,'' \emph{Proc. of the 26th International Technical Meeting of the
  Satellite Division of The Institute of Navigation (ION GNSS+ 2013)}, pp.
  1207--1214, 9 2013.

\bibitem{Kingma2014AdamAM}
D.~P. Kingma and J.~Ba, ``Adam: A method for stochastic optimization,''
  \emph{CoRR}, vol. abs/1412.6980, 2014.

\bibitem{svm_imbanlence}
R.~Batuwita and V.~Palade, \emph{Class Imbalance Learning Methods for Support
  Vector Machines}.\hskip 1em plus 0.5em minus 0.4em\relax John Wiley \& Sons,
  Ltd, 2013, ch.~5, pp. 83--99.

\bibitem{dist_shift}
O.~Wiles, S.~Gowal, F.~Stimberg, S.-A. Rebuffi, I.~Ktena, K.~Di~Dvijotham, and
  A.~T. Cemgil, ``A fine-grained analysis on distribution shift,'' in
  \emph{International Conference of Learning Representations}, 2022.

\bibitem{bi_lstm}
Y.-L. He, L.~Chen, Y.~Gao, J.-H. Ma, Y.~Xu, and Q.-X. Zhu, ``Novel double-layer
  bidirectional {LSTM} network with improved attention mechanism for predicting
  energy consumption,'' \emph{ISA Transactions}, vol. 127, pp. 350--360, 2022.

\end{thebibliography}
\end{document}